# Design, modeling, and characteristics of ring-shaped robot actuated by functional fluid

Zebing Mao[1*], Xuehang Bai[2*], Yanhong Peng[3], Yayi Shen[4],

[1] Faculty of Engineering, Yamaguchi University, 2-16-1 Tokiwadai, Ube City, Yamaguchi, 755-8611, Japan

[2] Schaeffler Intelligent Driving Technology (Changsha) Co., Ltd., Building A5, Hunan Inspection Detection Characteristic Industrial Park, No.336 Xueshi Road, Yuelu District, Changsha, 410000, Hunan, China

[3] Department of Information and Communication Engineering, Graduate School of Engineering, Nagoya University, Nagoya 4648601, Japan

[4]Department of Mechanical engineering, Nanjing University of Aeronautics and Astronautics, China

\* These authors contributed equally to this work.
Corresponding author: mao.z.aa@yamaguchi-u.ac.jp

**Abstract**

The controlled actuation of hydraulic and pneumatic actuators has unveiled fresh and thrilling opportunities for designing mobile robots with adaptable structures. Previously reported rolling robots, which were powered by fluidic systems, often relied on complex principles, cumbersome pump and valve systems, and intricate control strategies, limiting their applicability in other fields. In this investigation, we employed a distinct category of functional fluid identified as Electrohydrodynamic (EHD) fluid, serving as the pivotal element within the ring-shaped actuator. A short stream of functional fluid is placed within a fluidic channel and is then actuated by applying a direct current voltage aiming at shifting the center of mass of the robot and finally pushed the actuator to roll. We designed a ring-shaped fluidic robot, manufactured it using digital machining methods, and evaluated the robot's characteristics. Furthermore, we developed static and dynamic models to analyze the oscillation and rolling motion of the ring-shaped robots using the Lagrange method. This study is anticipated to contribute to the expansion of current research on EHD flexible actuators, enabling the realization of complex robotic systems.

## 1. Introduction

Mobile robots are anticipated to function in a variety of settings, such as performing routine safety monitoring tasks, identifying illnesses in diverse tiny organic creatures, optimizing the effectiveness of gathering natural resources, among other applications. In recent decades, there has been significant progress in creating mobile robots with diverse abilities for moving on land. In a general sense, these activities can be divided into four groups: crawling, rolling, walking, and jumping(Ng *et al.*, 2021; Ding *et al.*, 2023; Kong *et al.*, 2024). When compared to walking, crawling, and jumping robots, rolling robots offer several advantages: rolling robots equipped with rings, balls, or confined constructions offer shading to protect the inner components of the robot; rolling robots exhibit high adaptability when navigating unfamiliar terrains, making them versatile in challenging environments. Furthermore, rolling motion is a prevalent phenomenon observed in many animals and insects, enabling them to escape from perilous natural surroundings(Brackenbury, 1997).

So far, rolling robots with various rolling mechanism and actuating methods have been developed(Halme, Schonberg and Wang, 1996; Crossley, 2006; Floyd and Minor, 2010; Lin, Leisk and Trimmer, 2011; Landa and Pilat, 2015). These robots can be mainly grouped into four distinct categories: ordinary rolling, gravity-based rolling, impulse based rolling and ballistic rolling. Typically, in the ordinary rolling approach, robots move using the energy generated by tracks or wheels. Rolling based on gravity can be achieved by employing a ball, ring, or shell that undergoes internal weight fluctuations or changes in shape(Crossley, 2006). The gravity-based rolling robots have several advantages compared with other types of rolling robots. (a) Efficiency Through Simplicity. Gravity-based rolling robots leverage the simplicity of gravity as a driving force, eliminating the need for complex motor systems or intricate mechanical components. By relying on the fundamental force that governs our universe, these robots achieve a streamlined design, enhancing efficiency and reliability. (b) Energy Conservation. Unlike traditional motor-driven robots, gravity-based rolling robots tap into a sustainable energy source – gravity itself. This not only reduces the dependence on external power supplies but also contributes to prolonged operational periods, making them ideal for applications requiring extended mission durations. (c) Reduced Maintenance Requirements. Simplicity in design often translates to reduced maintenance needs. Gravity-based rolling robots, with their minimalistic approach, present a lower risk of mechanical failures and are inherently more robust, potentially lowering the overall cost of ownership. An example that is widely-recognized would be that a typical gravity based rolling robot alter its center of mass by adjusting the placement of component parts, adopting the same principle as a pendulum(Halme, Schonberg and Wang, 1996; Landa and Pilat, 2015). This arrangement utilizes interaction between the ground and the robots, resulting in the creation of a moment arm from the centroid. As a consequence of the normal force exerted, the robot experiences a continuous rolling motion when the driving torque surpasses the torque produced by friction. The impulse method, which is different from gravity based rolling system, involve utilizing an adjustable arm that extends from the robot's center to its outer region. The ballistic rolling approach employs a segmented rolling mechanism and anchoring components to achieve the expansion and contraction of the robot's movement(Lin, Leisk and Trimmer, 2011). Nonetheless, a number of these systems can be unwieldy and bulky because of inflexible elements such as motors, pendulums, gears, and so on.

Soft rolling robots, with the application of recent progress in soft materials and actuators, have the advantage of not requiring any rigid components. These soft materials simplified the intricate structural design and control strategies involved, which provide high degree of freedom and lower weight (Kong *et al.*, 2023; Peng *et al.*, 2023b; Peng *et al.*, 2024). Contrary to inflexible rolling robots, the key characteristics of soft rolling robots are their ability to change shapes and quickly respond to external stimuli to shift their center of mass. Ding et al. considering the compressibility of the hydraulic cavity medium, proposed an optimal trajectory planning method based on vibration polynomials, which can effectively reduce the vibration of mobile pumping equipment(Ding *et al.*, 2024). Based on photo mobile polymer materials, Yamada *et al.* (2008) managed to create light-driven soft motors which convert light energy into mechanical work. Photo motors, consisting of a single-walled carbon nanotube (SWNT) /polymer bilayers, have been researched and manufactured by Zhang *et al.* (2014). However, one of limitations is that these photomechanical robots usually produce restricted speed. Hu *et al.* (2018) developed a soft robot that achieves high-speed locomotion through shape-changing mechanisms induced by an applied magnetic field. This method enables the manipulation and guidance of the robot along different routes using the manipulation of magnetic fields. This necessitates a complex system and substantial equipment for its implementation. Wu *et al.* (2018) developed a wheeled robot driven by a liquid-metal droplet, investigating a method to enable locomotion outside a liquid environment using an enclosed system as the core of the driving module. A rolling robot of star shape is developed by Huang *et al.* (2018) who employed seven curved SMA (shape memory alloy) actuators. Sugiyama *et al.* (2005) suggested a circular robot that is capable of achieving deformation in its body based on the utilization of morphing SMA coils. Dielectric elastomer actuators (DEAs) are utilized to create rolling robots, benefiting from their lightweight nature and rapid responsiveness(Potz *et al.*, 2010; Li *et al.*, 2021). Potz *et al.* (2010) proposed a pneumatic silicone hollow cylinder comprising segments of dielectric elastomer actuators. Li *et al.* (2021) researched and manufactured three distinct rolling robots, each employing unique mechanisms, including gravity-based rolling, impulse-based rolling, and ballistic rolling. Besides these methods, fluidic systems are one of the candidates to developing the rolling robots (Fang *et al.*, 2021; Jiao *et al.*, 2023a; Jiao *et* *al.*, 2023b). Gong (2015) devised a distinctive system utilizing the peristaltic movement of a stator and rotor made of elastic material. Preston *et al.* (2019) combined a ring oscillator with a rolling robot with a hexagonal shape. Robertson and Paik (2017) presented a soft pneumatic actuator driven by vacuum to achieve a rolling moving mode. The actuator features multiple degrees of freedom (DOFs). Nevertheless, within these dynamic systems, the rolling mechanism operates on impulses, utilizing a deformed chamber to facilitate the rolling procedure. As a result, extended cables become essential, and control logic for each actuator are intricate.

Aiming to overcome these limitations, we propose an EHD pump, which is a steady stream of non-conductive liquids within the channel. The proposed EHD pump solution addresses limitations in existing rolling robots by enabling continuous rolling movement without impulse-based actuators, simplifying control strategies with voltage dependence, and featuring a lightweight, flexible structure, thereby overcoming obstacles in development and practical implementation. Fig. 1 shows that the flow jet is induced via the application of high DC voltage. EHD pumps with inflexible structure were proposed and developed a few decades ago(Bryan and Seyed-Yagoobi, 1991), which consists of the rigid base, rigid cover, and metal electrodes. The EHD phenomenon has been applied to microfluidics(Mao, Yoshida and Kim, 2019a; Mao, Yoshida and Kim, 2019b; Matsubara *et al.*, 2022) and soft robots(Mao *et al.*, 2020; Mao, Iizuka and Maeda, 2021) etc. The characters of EHD pumps have been studied experimentally or based on the machine learning method( Mao *et al.*, 2023; Zhang *et al.*, 2023). In the literature we previously reported, rolling robots were required to operate in a liquid environment due to electrostatic forces from the ground, using bands to maintain their shapes (Mao *et al.*, 2022). These robots also demonstrated slow rolling speeds. To address these limitations, we have adopted a simple design that incorporates a soft-structured EHD pump to enable continuous rolling movement. The controlling strategies are simple and only depend on the applied voltages. The EHD pump revolutionizes rolling robots, enabling continuous movement through oscillating fluids. Its simple control strategies, based on applied voltages, streamline logic compared to alternatives. With a lightweight, flexible structure devoid of rigid components, it simplifies design, providing freedom for shape changes. Compatibility with soft materials leverages recent progress, simplifying both design and control strategies. The EHD pump's potential for higher speeds marks a significant progression in rolling robot technology.

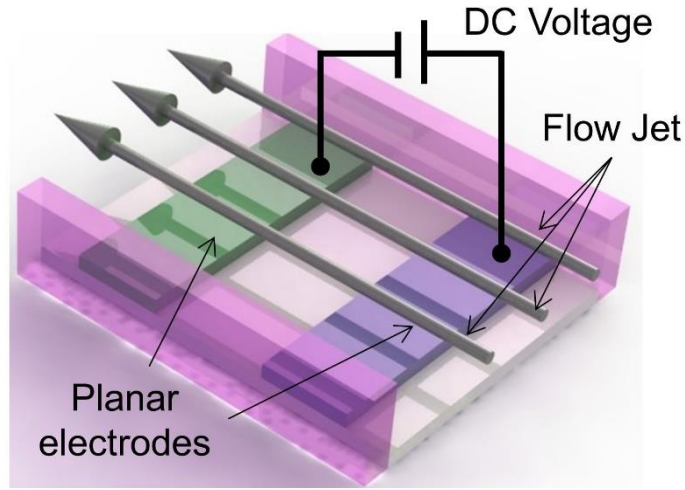

**Fig. 1** Schematics of EHD pump

## 2. Principle, design, and modeling
### 2.1 Principle and design

In line with prior literature, it is well-established that EHD pumps are typically filled with specialized liquids to prevent voltage breakdown(Peng *et al.*, 2023a). In our research, we take advantage of situations in which both EHD liquid and air are present in the fluidic channel to develop a EHD roller of soft structure. For the purpose of initiating the motion of the roller, a precise amount of EHD liquid was pumped into the fluidic channel before assembly. In previous research, EHD pumps typically required the presence of EHD liquids to enable locomotion. Commonly used liquids for EHD pumps include hydrocarbons, fluorinert compounds, and hydrofluoroethers (HFEs), such as HFE 7100, dibutyl decanedioate (DBD), and electronic liquid FC-40 etc. Among these options, HFEs are favored due to their superior thermal and chemical stability and their dielectric properties, which resemble those of perfluorinated fluids. Consequently, they are frequently chosen as the preferred EHD pump liquids.

Additionally, HFEs offer higher power generation capabilities compared to fluorient-based liquids and exhibit relatively high compatibility with materials, including plasticizers, in contrast to hydrocarbon-based liquids. In our specific case, we utilize HFE 7300 (3M™ Novec™ Engineered Fluids) for our EHD pumps. The EHD ring incorporates two EHD pumps, providing both twistability and bendability. This flexibility significantly enhances the robot's overall functionality and expands its potential applications. It enables the robot to navigate intricate environments, execute diverse tasks, and adapt to various scenarios, as illustrated in Fig. 1. To activate the EHD ring, we introduce 0.4 ml of HFE 7300 into the EHD ring. The underlying principle is succinctly articulated within this discourse: Two EHD pumps are integrated into a single EHD ring. When no direct current voltage is applied, the HFE liquid resides at the bottom of the EHD ring (**Fig. 2**a). The EHD fluid rises when voltage is applied (**Fig. 2**b), generating a driving torque due to the moment arm provided by the distance between the center of the EHD ring and the point of interest. Nevertheless, the torque produced is initially insufficient, mainly due to the lightweight nature of the EHD fluid, and struggles to overcome the friction torque (**Fig. 2**b). At this point, we briefly deactivate the applied voltage, causing the EHD fluid to return to its original position (**Fig. 2**c). To facilitate EHD fluid pumping, we then re-apply the voltage (**Fig. 2**d), thereby increasing the driving torque. The design features elevated electrodes, which significantly reduces the electrostatic interaction with the ground. This innovation allows our robot to operate in a dry environment, eliminating the necessity for a liquid medium. This system allows the EHD ring to move across the surface.

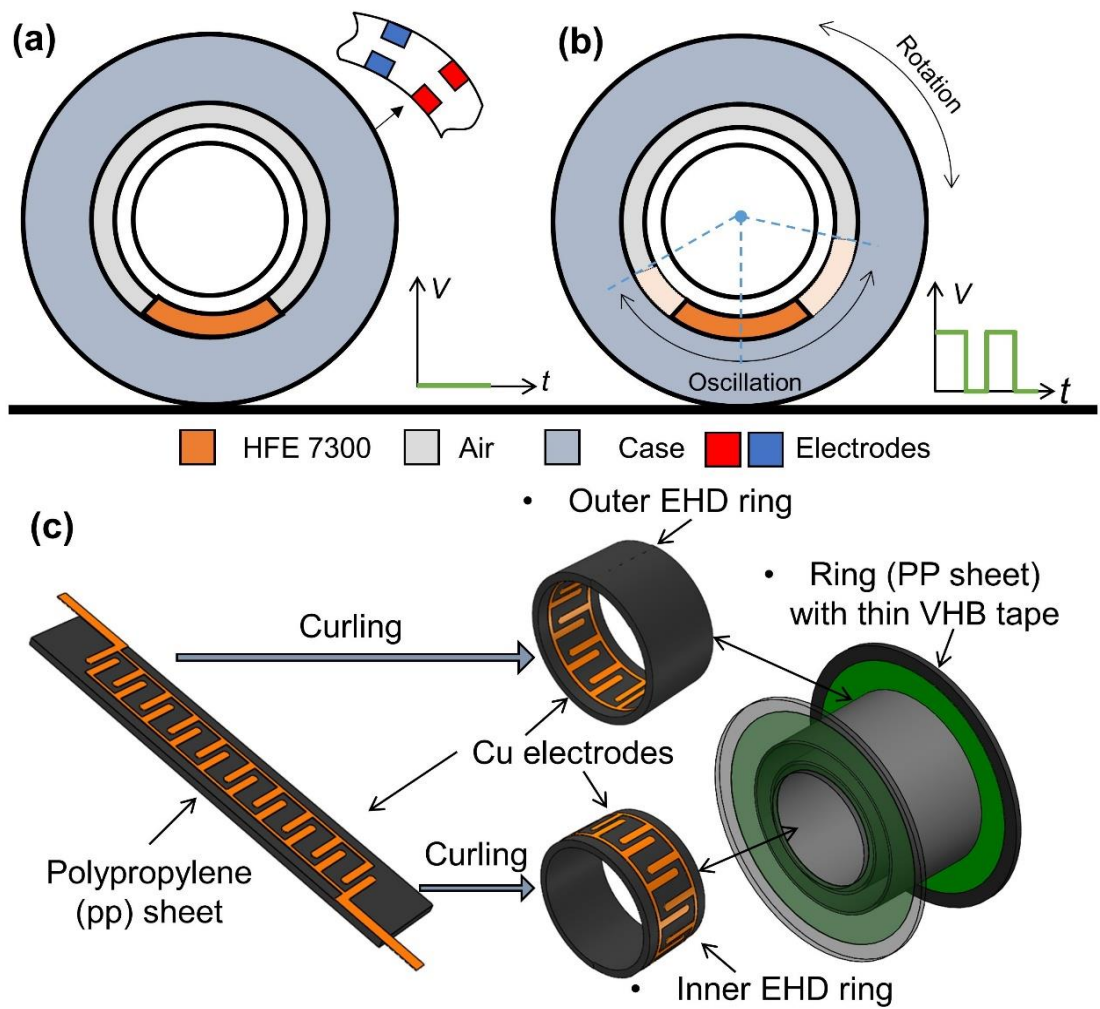

**Fig. 2** Working principle and design of EHD ring-shaped robot. (a) Initial state. (b) Rotation generation by the application of rectangular wave to the robot that resulting in fluidic oscillation. (c) Design of the EHD ring using 3D modeling method.

**2.2 Simulation of EHD pumps**

We used the physical simulation software, namely, COMSOL Multiphysics 5.6 (COMSOL Inc.), to predict the pressure and rate of flow that our soft EHD pumps generates (**Fig. 3**). The procedures involved a two-step approach: (a) Utilizing electrical field modeling to obtain a simulated pressure, and (b) Implementing the finite element method (FEM) to determine the simulated deflection of the bending finger by applying the obtained pressure. In **Fig. 3**a, we model the electric field, followed by determining the volume force. Ultimately, we derive the resulting pressure and flow rate, considering the EHD flow as laminar flow. The Korteweg-Helmholtz equation expresses the electric body force exerted on the fluid(Felici, 1972):

$$\vec{F} = q\vec{E} - \frac{\varepsilon_0}{2}E^2\nabla\varepsilon_r + \frac{\varepsilon_0}{2}\nabla(E^2\frac{\partial \varepsilon_r}{\partial \rho}\rho) \qquad (1)$$

where the permittivity of vacuum is denoted as $\varepsilon_0$, the relative permittivity of the liquids is denoted as $\varepsilon_r$, electric field is denoted as $\vec{E}$, charge density is denoted as $q$ and density of the liquid is denoted as $\rho$. Within the context of the equation (4), the Coulomb force, dielectric force, and electrostriction force correspond to the first, second, and third terms, respectively. Of all these influences, the Coulomb force plays the most significant role in generating the flow. By the expanding Gauss's law and the Laplace operator, we can derive:

$$q = \nabla \cdot (\varepsilon_0 \varepsilon_r \vec{E}) \qquad (2)$$

The relationship below is obtained by assuming a stationary condition and constant electrical permittivity ($\varepsilon$):

$$\nabla \cdot J = 0 \qquad (3)$$

To simplify, the electric current density (*J*) is related to the electric field and the electric conductivity ($\sigma$) by Ohm's law :

$$J = \sigma \vec{E} \qquad (4)$$

By substituting the equation (5) and equation (7) to the equation (4) and equation (6), respectively, we can obtain the necessary equations for modeling the electrical field:

$$\vec{F} = \vec{E}\nabla \cdot (\varepsilon_0 \varepsilon_r \vec{E}) \qquad (5)$$

$$\nabla \cdot (\sigma \vec{E}) = 0 \tag{6}$$

The magnitude of the electric field can be obtained by the AC/DC module (electric currents and laminar flow interfaces) from COMSOL Multiphysics when we designed the electrode patterns. the flow rate is given by using the electric force according to the Navier-stokes equations:

$$\frac{\partial \vec{\mu}}{\partial t} = -(\vec{\mu} \cdot \nabla)\vec{\mu} - \frac{1}{\rho}\nabla p + \frac{\eta}{\rho}\nabla^2 \vec{\mu} + \frac{1}{\rho}q\vec{E} \tag{7}$$

where $\rho$, $\vec{\mu}$, $p$ and $\eta$ are the density (1660 kg/m$^3$), the flow rate, pressure, and viscosity (1.18 mPa·s), respectively. The relative permittivity and electric conductivity of the medium in our study are 6.1 and 2.1×10$^{-9}$ S/m. This process is completed by the laminar flow module from COMSOL Multiphysics. In our process, we only consider one pair of copper electrodes and then use its pressure (flow rate) to multiply the quantities of electrodes, which can calculate the overall output of the flexible EHD pump. Since our EHD pumps are flexible, the modeling process should contain the electrical field, flow field, and geometrical deformation. However, the geometrical deformation is caused by pressure loss and fluid structural interaction, which make it difficult to simulate. In this case, we introduce the empirical coefficients of flow rate ($C_q$) and pressure ($C_p$) to amend the simulation models. We simulated the EHD pumps with their applied voltage increasing 1 kV to 11 kV gradually. To verify the values of $C_p$ and $C_q$, we designed and fabricated a normal pump with its parameters (**Fig. 3**b) and then tested the generated pressure and flow rate. We could easily derive the values of $C_p$ and $C_q$ from the results in **Fig. 3**c and **Fig. 3**d, which are 15 and 11, respectively. For sake of simplicity, we also curve-fitted and retract the function which is analogous to **Fig. 3**c and **Fig. 3**d. The experimental setup is described in the supplement file. Therefore, the functions in terms of pressure and flow rate can be given by:

$$P = 2.152V^2 - 2.031V \tag{8}$$

$$Q = 0.0076V^2 + 0.0167V \tag{9}$$

where $P$, $Q$ and $V$ are generated pressure, flow rate generated by EHD pumps and applied voltages, respectively.

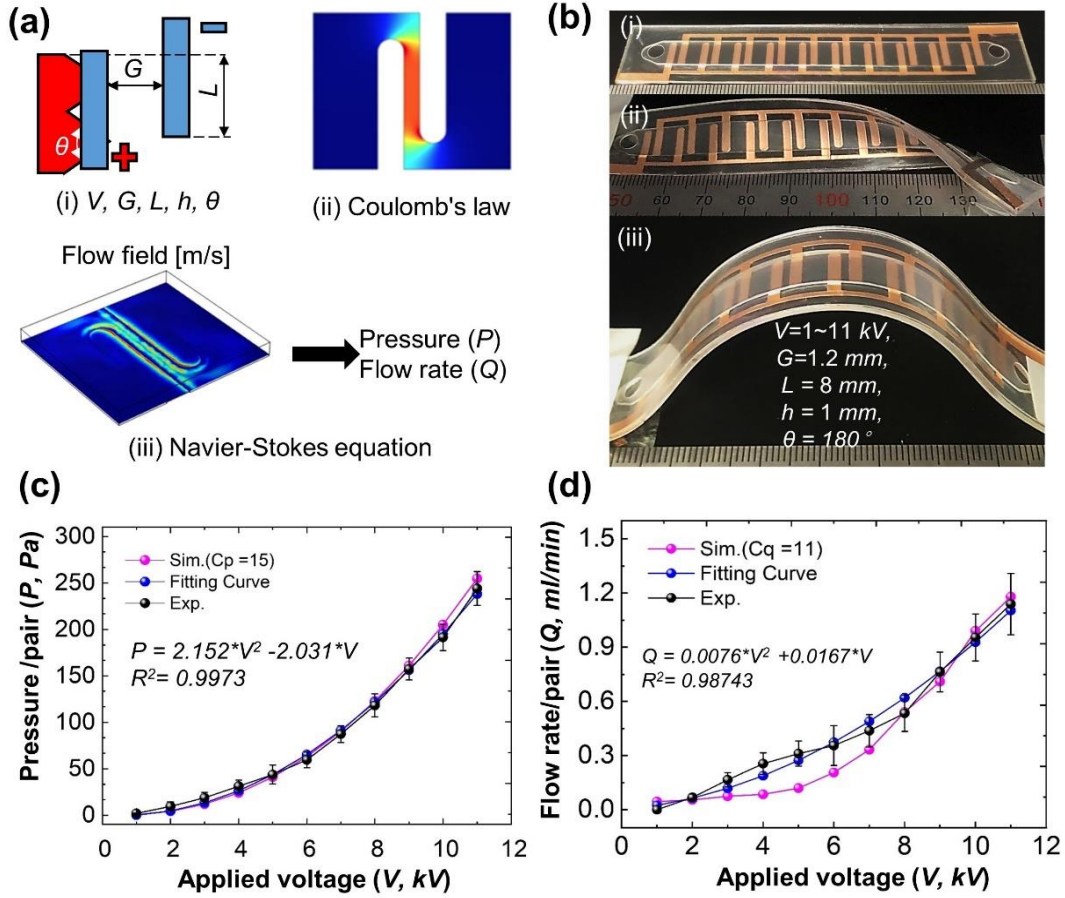

**Fig. 3** (a) Simulation process, (i) schematic diagram of electrode: V: Voltage; G: Electrode spacing; L: Length of the electrodes; h: Height of the channel; θ: Angle of inclination of the electrodes, (ii) application of Coulomb's law, (iii) Navier Stokes equation diagram (b) Normal pumps, (i) straight state, (ii) twisted state, (iii) bending state; (c) Pressure performance, values of $C_p$ and seeking the pressure-related function. (d) Flow rate performance, values of $C_q$ and seeking the flow rate-related function.

## 2.3 Modeling

**Start to roll:**

For the purpose of modeling this setup, we consider the mass, rotational speed, and moment of inertia of the ring, as well as the EHD fluid, denoted as $m_1$ and $m_2$, $\dot{\theta}_1$ and $\dot{\theta}_2$, $J_1$ and $J_2$ respectively (**Fig. 4**a). We define the outer and inner radius of the EHD ring to $R_1$ and $R_2$. $g$ represents the acceleration of gravity. At this stage, the ring-shaped robot starts to roll and have to overcome the friction and therefore, the static configuration (with a maximum possible driving torque) is: $\theta_1 = 0$, $\theta_2 - \theta_1 = \theta_{max} =$ $\theta$. Considering the static model, to drive the ring-shaped robot, we define the possible maximum driving torque as $T_{drive}$.

$$T_{drive} = m_2 g R_2 \sin\theta \tag{10}$$

To induce rotational movement, the EHD rolling robot needs to possess a torque of the highest achievable magnitude to counteract the torque generated by rolling friction. The resistance force arising from the interaction between the rolling robot and the ground surface can be expressed as:

$$f_{friction} = k_1(m_1 + m_2)g \tag{11}$$

where $k_1$ presents coefficient of friction (COF). The driving torque applied can be expressed as:

$$T_{drive} - k_1 R_1 (m_1 + m_2) g f_{Sign(\omega_1)} = J_1 \ddot{\theta} \tag{12}$$

where $T_{drive}, \ddot{\theta}$ are the driving torque. The angular acceleration of the rolling ring, denoted as $f_{Sign(\omega_1)}$, can be expressed as:

$$f_{Sign(\omega_1)} = \begin{cases} 1 & \omega_1 > 0 \\ 0 & \omega_1 = 0 \\ -1 & \omega_1 < 0 \end{cases} \tag{13}$$

Besides, if we focus on the highest point in the ring where the EHD fluid is most concentrated, we can express the connection between µ and θ using the following equation, excluding the effects of sliding friction:

$$0.5 m_2 \mu^2 = m_2 g(1-\cos\theta) \tag{14}$$

where, $\mu$ is the average velocity of EHD fluid. If we take the cross-section area of the fluidic channel ($A$) into account, this value can be obtained by:

$$Q = A\mu \tag{15}$$

Here, we consider the applied voltage with the shape of the square wave, which can be described by:

$$V = DV_{max} + (1 - D)V_{min} \tag{16}$$

where $D \in (0\sim1)$, $V_{max}$, $V_{min}$ are the duty, maximium and minmiun voltages, respectively. In our case, $V_{min}$ equals to $0\ kV$. Therefore, the angular velocity is described as follows:

$$\dot{\theta} = DT/J_1 \left\{ m_2 g R_2 \sin\left(arc\left(1 - \frac{(0.0076V^2 + 0.0167V)^2}{2A^2 g}\right)\right) - k_1 R_1 (m_1 + m_2) g \right\} \tag{17}$$

**Dynamic model:**

We created a model that captures the dynamic nature of rolling on a flat surface using the Lagrange method to gain a deeper understanding of the locomotion of our ring-shaped robot. Before introducing an external voltage, the ring-shaped robot remains stationary on the flat surface, with the contact point O as the pivot. Upon activation, the functional fluid exerts a driving force, which can be derived by:

$$F = PA \tag{18}$$

This force initiates the motion of the module responsible for propelling the wheel along its flanges, setting the ring-shaped robot into motion due to the imbalance in its center of gravity (CG). After a short time, point O relocates to a new position, and the wheel's body undergoes a rotation of $\theta_1$ degrees, while the functional liquid rotates by $\theta_2$ degrees. Therefore, according to the Lagrange equations, the system is given by(Landa and Pilat, 2015):

$$L_o = K_{ring} + K_{EHD} + T_{ring} + T_{EHD} - U_{ring} - U_{EHD} \tag{19}$$

where $K_{ring}$, $K_{EHD}$, $T_{ring}$, $T_{EHD}$, $U_{ring}$, $U_{EHD}$ are kinetic energy, rotational energy, potential energy of the EHD ring and EHD fluid, respectively. The referred point of potential energy is defined as the ring's centroid. Kinetic, rotational and potential, energies are given by:

$$K_{ring} = 0.5 m_1 (\dot{\theta}_1 R_1)^2 \tag{20}$$

$$K_{EHD} = 0.5 m_2 (\dot{\theta}_1 R_1)^2 + m_2 R_1 R_2 (\dot{\theta}_2 - \dot{\theta}_1) \cos(\theta_2 - \theta_1) \tag{21}$$

$$T_{ring} = 0.5 J_1 (\dot{\theta}_1)^2 \tag{22}$$

$$T_{EHD} = 0.5 J_2 (\dot{\theta}_2 - \dot{\theta}_1)^2 \tag{23}$$

$$U_{ring} = m_1 g R_1 \tag{24}$$

$$U_{EHD} = m_2 g R_2 [1 - \cos(\theta_2 - \theta_1)] \tag{25}$$

We can obtain the Rayleigh dissipative, which is given by:

$$\psi = 0.5 \left( \xi_{m1} (\dot{\theta}_1)^2 + \xi_{m2} (\dot{\theta}_2)^2 \right) \tag{26}$$

where $\xi_{m1}$ and $\xi_{m2}$ denotes the viscous coefficients of friction between the ring /functional liquid and the fluidic channel, respectively. Considering the generalized coordinate of the ring-shaped robot and the force of the system, we can also derive Lagrange equation with Rayleigh dissipative:

$$\frac{d}{dt} \frac{\partial L_o}{\partial \dot{\theta}_1} - \frac{\partial L_o}{\partial \theta_1} + \frac{\partial \psi}{\partial \dot{\theta}_1} = 0 \tag{27}$$

$$\frac{d}{dt} \frac{\partial L_o}{\partial \dot{\theta}_2} - \frac{\partial L_o}{\partial \theta_2} + \frac{\partial \psi}{\partial \dot{\theta}_2} = F R_2 \tag{28}$$

Therefore, linear rolling equation of the ring-shaped robot as follows:

$$M \begin{bmatrix} \ddot{\theta}_1 \\ \ddot{\theta}_2 \end{bmatrix} + K \begin{bmatrix} \dot{\theta}_1 \\ \dot{\theta}_2 \end{bmatrix} = C \tag{29}$$

where:

$$M = \begin{bmatrix} J_1 + m_1 R_1^2 + m_2 R_1^2 + J_2 - 2m_2 R_1 R_2 \cos(\theta_2 - \theta_1) & -J_2 + m_2 R_1 R_2 \cos(\theta_2 - \theta_1) \\ m_2 R_1 R_2 \cos(\theta_2 - \theta_1) - m_2 R_2^2 & m_2 R_2^2 \end{bmatrix}$$

$$K = \begin{bmatrix} m_2 R_1 R_2 (\dot{\theta}_2 - \dot{\theta}_1) \sin(\theta_2 - \theta_1) + \xi_{m1} & m_2 R_1 R_2 (\dot{\theta}_2 - \dot{\theta}_1) \sin(\theta_2 - \theta_1) + \xi_{m1} \\ 0 & \xi_{m2} \end{bmatrix}$$

$$C = \begin{bmatrix} m_2 g R_2 \sin(\theta_2 - \theta_1) \\ (2.152 V^2 - 2.031 V) A R_2 - m_2 g R_2 \sin(\theta_2 - \theta_1) \end{bmatrix}$$

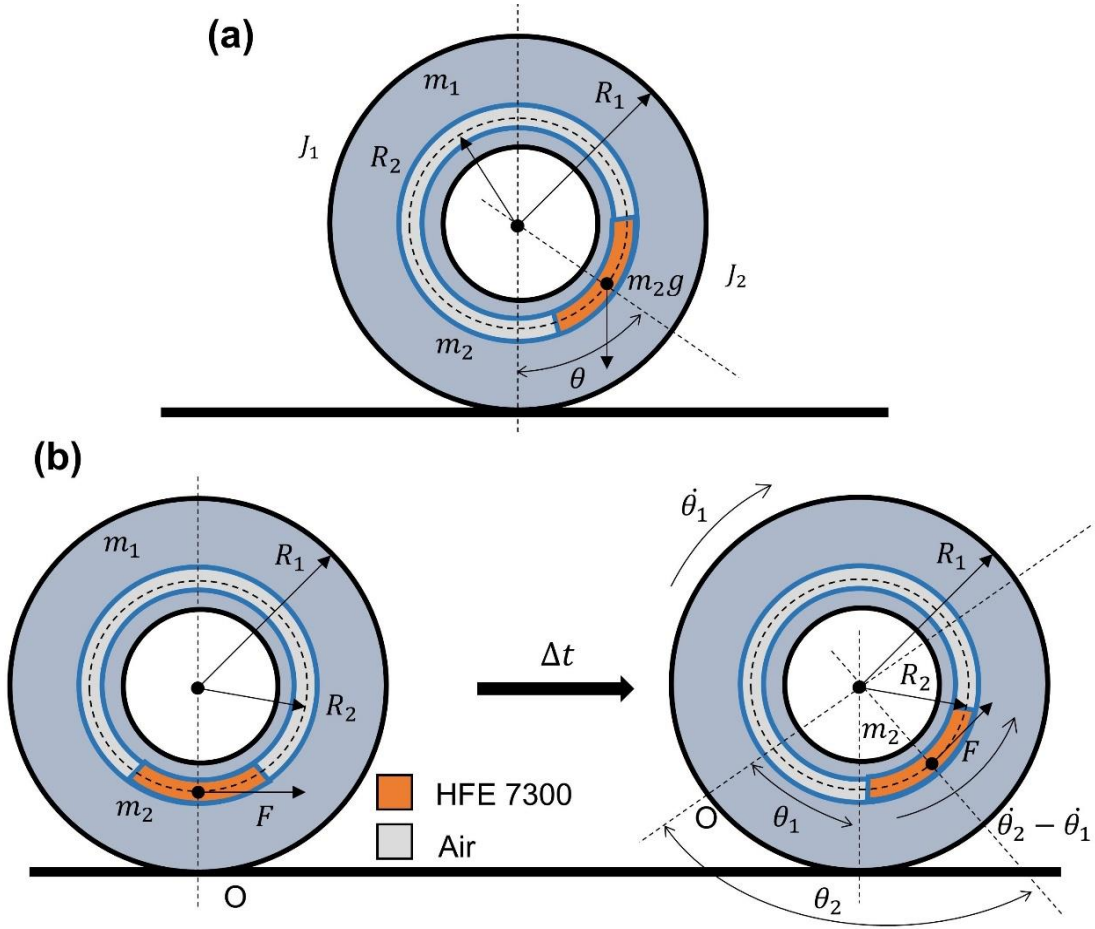

**Fig. 4** Simplified model of ring-shaped robot actuated by functional fluid. (a) Start to roll. (b) Dynamic model of the robot

## 3. Experiment, Results and Discussion

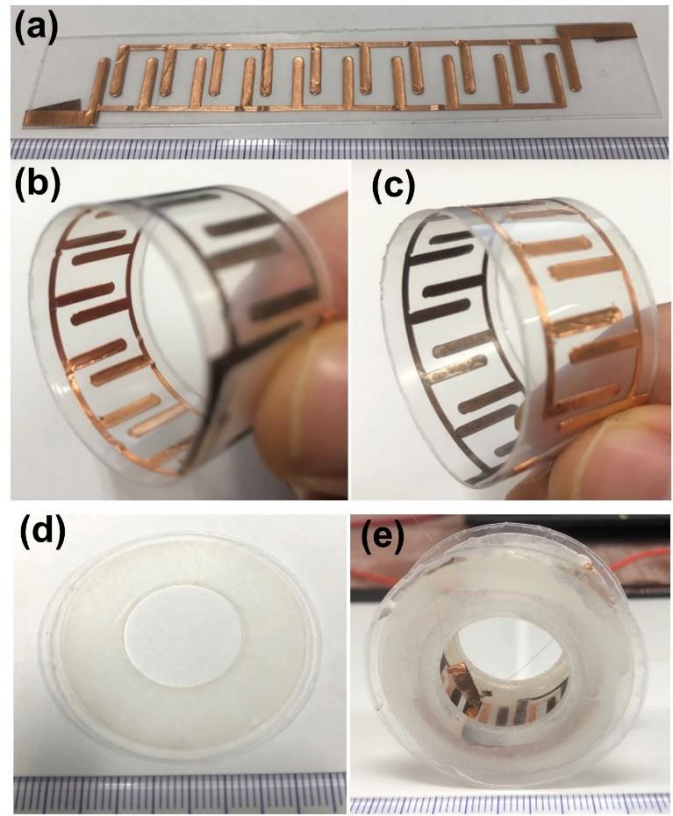

**Fig. 5** Manufacture process of the EHD ring. (a) Normal EHD pump. (b) Outer curled EHD pump. (c) Inner curled EHD pump. (d) Wheeled ring. (e) Assembled devices.

The EHD pump consists of a dual layered structure, comprising a layer made of polypropylene (PP) and electrodes crafted from copper. The EHD pump can be bent and twisted because of the flexible and pliable characteristics of the thin PP layer. The outer and inner EHD rings have ten and nine pairs of Cu electrodes. Considering the design, the two electrodes had an overlapping distance of 8 mm and a gap of 1.2 mm between them, respectively. The dimensions of a single electrode were 10 millimeters in length, 2 millimeters in width, and 35 micrometers in height. We positioned the two sets of electrodes with a 4 mm gap between them. Our manufacture is based on the digitally fabricated method. Briefly, a sheet of Cu was initially attached to the prepared PP sheet. The adhesive layer between the Cu layer and the PP sheet was 35 μm thick. Then we used a cutting plotter (GRAPHTEC CE6000-40 Plus) to fabricate the interdigitated electrodes. After stripping the unnecessary parts of the Cu sheet, we separated each EHD pump via a laser cutter (TROTEC Speedy 100). The wheel ring and the thin VHB layer are also fabrciated by using the laser cutter. Finally, we assembled the components to form an EHD ring (**Fig. 5**).

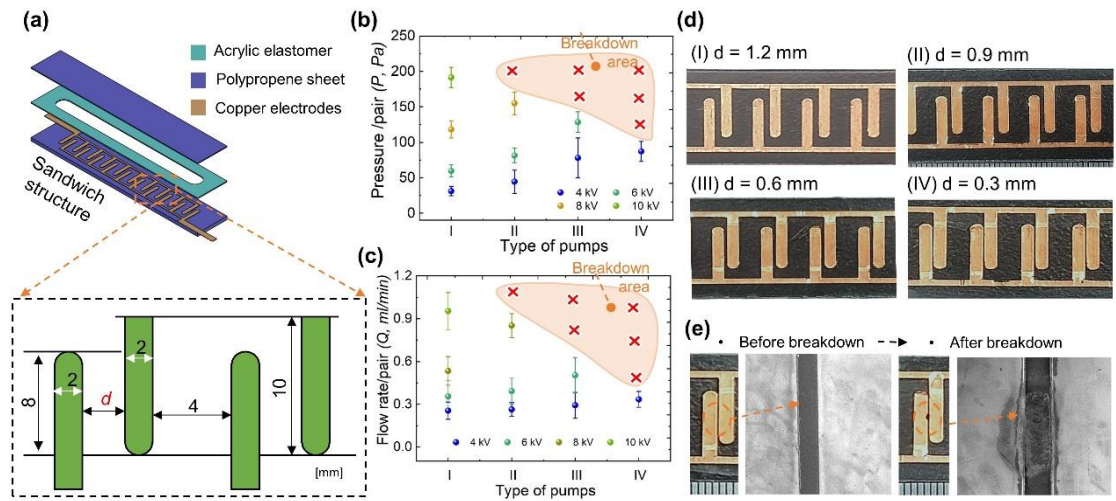

**Fig. 6** Impact of Electrode Spacing on Pressure and Flow Rate under Variable Voltage Conditions (a) Size of Sandwich-structured electrode (b) Pressure influenced by pump type (c) Flow rate influenced by pump type (d) Four types of Electrode arrangement (e) Pictures of Electrode breakdown.

Regarding the selection of electrode spacing, **Fig. 6** illustrates that closer electrode spacing correlates with higher pressure and flow rate at identical voltage levels. However, this correlation becomes precarious under high voltage conditions, as excessive proximity may result in voltage breakdown and subsequent device failure. From **Fig. 6**(b), electrode configurations with spacings of 0.9mm and 0.6mm experience breakdown at 10kV and 8kV respectively. This risk intensifies with narrower spacings, with breakdown occurring at 6kV, 8kV, and 10kV for a 0.3mm spacing. Consequently, this phenomenon presents a significant operational challenge under high voltage conditions. To mitigate this risk, we have selected an electrode spacing of 1.2mm, capable of withstanding 10kV. While this choice does introduce drawbacks such as reduced pressure and flow rate at the same voltage, it enables operation at higher voltages without encountering electrode breakdown. Thus, the selection of 1.2mm electrode spacing facilitates greater pressure generation under increased voltage application, ensuring operational integrity.

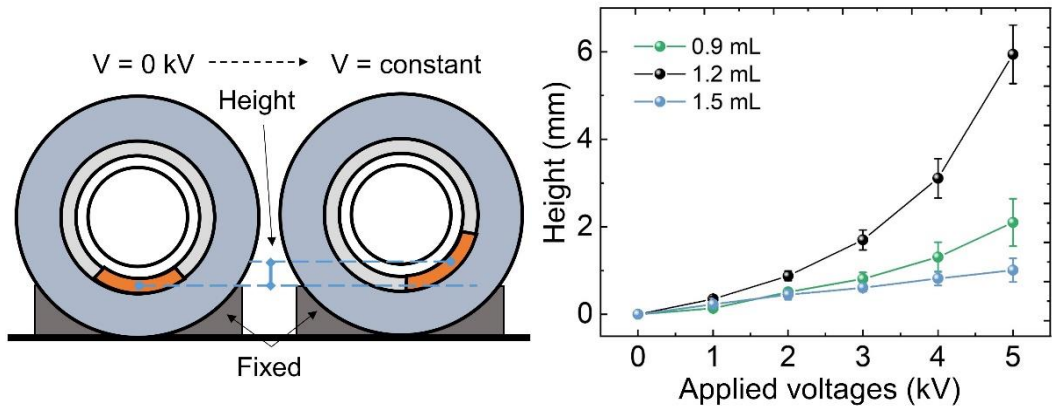

**Fig. 7** Effect of Liquid Volume on Lifting Height Under Applied Voltage

For the investigation of liquid volumes, we conducted additional experiments, the results of which are illustrated in the **Fig. 7**. Specifically, we introduced liquid volumes of 0.9ml, 1.2ml, and 1.5ml into the inner cavity, respectively, while applying voltages ranging from 0 to 5kV. As depicted in the left-hand image, the application of voltage elevates the center of mass of the liquid. The corresponding outcomes are portrayed in the right-hand image. Notably, the 1.2ml volume exhibits the greatest elevation under identical voltage conditions. This observation can be attributed to the fact that insufficient liquid diminishes the number of electrode pairs in contact, thereby reducing the influence of the electric field. Conversely, excessive liquid leads to an increase in the gravitational potential energy required for upward movement, thereby limiting the lifting height. Through experimentation, we determined that 1.2ml represents the optimal liquid volume for injection.

In this experimental investigation, the primary focus was on the design and fabrication of a compact high-voltage circuit for the EHD ring-shaped robot. The circuit, interfaced with a personal computer and Arduino, aimed to achieve specific voltage outputs through the use of a DC/DC converter. Subsequently, the study explored the nuanced relationship between applied electrical voltages and the resulting motion of the soft ring-shaped robot. The investigation also involved validating a proposed theoretical dynamic model by comparing its predictions with experimental observations of angular velocity. Notably, hypotheses were formulated, positing that controlled voltage and frequency manipulation, along with optimal duty cycles, would enable the desired actuation and consistent rotation of the robot under specific EHD pump activation conditions.

We designed and fabricated a compact high-voltage circuit for the EHD ring-shaped robot (**Fig. 8** & video S1). In essence, the amplifier is linked to a personal computer (PC)/ a power supply battery (9.0 V), and we employed an Arduino (5.0 V) to provide power to the direct current (DC/DC) converter (EMCO Q101-5) via a potentiometer. The potentiometer is employed for fine-tuning the output voltage. The DC voltage can be converted into a high direct current voltage through the use of a DC/DC converter. To output on-off voltage with a certain level of frequency, we utilized high voltage optocoupler (OPTO-150, HVM Technology, Inc.) and an Arduino Micro pulse-width modulation (PWM) for the creation of a low DC voltage characterized by a square wave. Since the optocoupler can withstand the high voltage over 10 kV, the switching process by the PWM signal can be transferred to the output of the DC/DC converter. Then, we manufactured the circuit using the substrate processing machine (KitMill CIP100, ORIGINALMIND Inc.). The total size of this power source are 43.21 g (without battery) and a dimension of $58 \times 53 \times 23$ mm$^3$, respectively (**Fig. 8**b). Our experimental findings demonstrated that the electrical circuit achieved the highest frequency of output of 20 Hz once the voltage output of the circuit reached 5.2 kV (**Fig. 8**c & **Fig. 8**d).

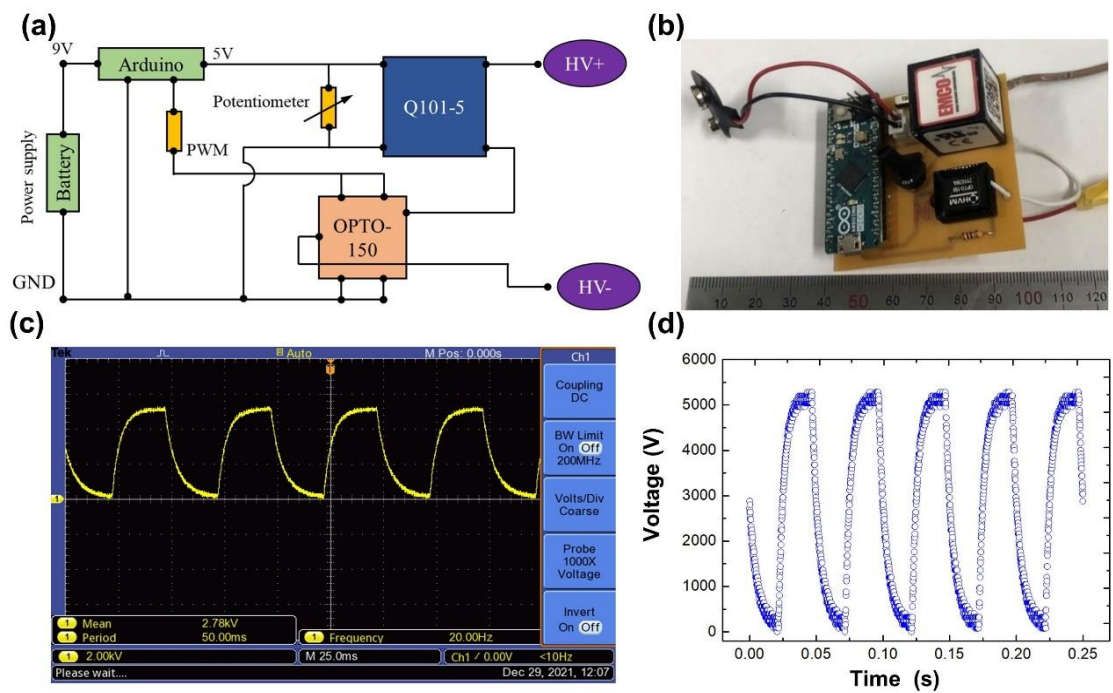

**Fig. 8** (a) Design of high voltage circuit. (b) Photos of the fabricated circuit. (c) Output voltage and frequency obtained by the oscilloscope. (d) Output voltage versus time of high voltage circuit.

We used the developed circuit to actuate the soft ring-shaped robot. The applied voltages were adjusted according to the experimental details. Initially, we explored the connection between the rotational speed and the effect of the applied electrical voltages, which is also relating to the boundary for the robot to start rolling (**Fig. 9**a). It can be found that the angular velocity increases and then reduces after the 70% duty. We also utilized the model to estimate this tendency, which follows with the experimental results. This is because the higher duty can cause the liquid to climb up along the fluidic channel for a longer time, which results in longer lever arm and therefore causing the higher driving torque. It is noted that the angular velocity reduces after 70% is because only amounts of liquid cannot directly generates enough torque to drive the roller. By using the voltage with appropriate duty, the swaying of the roller can finally move on the ground. In **Fig. 9**b, we could find that the ring-shaped robot oscillates with time goes by according to the position the robot stands. The applied voltages and frequencies to the rolling robot are 4.5 kV and 5 Hz, respectively (**Fig. 9**a&b). Therefore, we choose the duty of 70% to drive the robot in the following robot.

Our experiments reveal that the angular velocity increases with the time (**Fig. 9**c& video S2), which can be divided into a phase of increasing speed and a phase of consistent rolling. In its default state, prior to the application of an external electrical voltage, the ring-shaped robot maintains a static position on the flat surface. Upon the initiation of the electrical voltage, the operational fluid rapidly induces a propulsive force. This force activates the module tasked with propelling the wheel, which is mounted along its flanges, thereby inducing movement of the ring-shaped robot due to an alteration in the distribution of its center of gravity. Our simulating parameters can be found in the supplementary file. As long as the EHD pump is activated under the voltage of 4.5 kV, frequency of 5 Hz and duty of 70%, the functional liquid moves along the fluidic channel and caused the robot to maintain a consistent pace of rotation with response time less than 1 s. The ring-shaped robot begins at a high angular acceleration, which induce it to reach a steady speed (0.8 rad/s) in a short time. We also calculate the angular velocity based on our proposed theoretical dynamic model, which agrees with the experimental results. The rolling motion of the ring-shaped robot can be found in **Fig. 9**d, in which it starts at the initial position of 1.9 mm (ruler) and ends at the 5.3 mm within less than 3 s. The robot is ability to roll continuously in a forward direction when we set the applied voltage of 5.5 kV, a frequency of 5 Hz and duty of 70% (video S3).

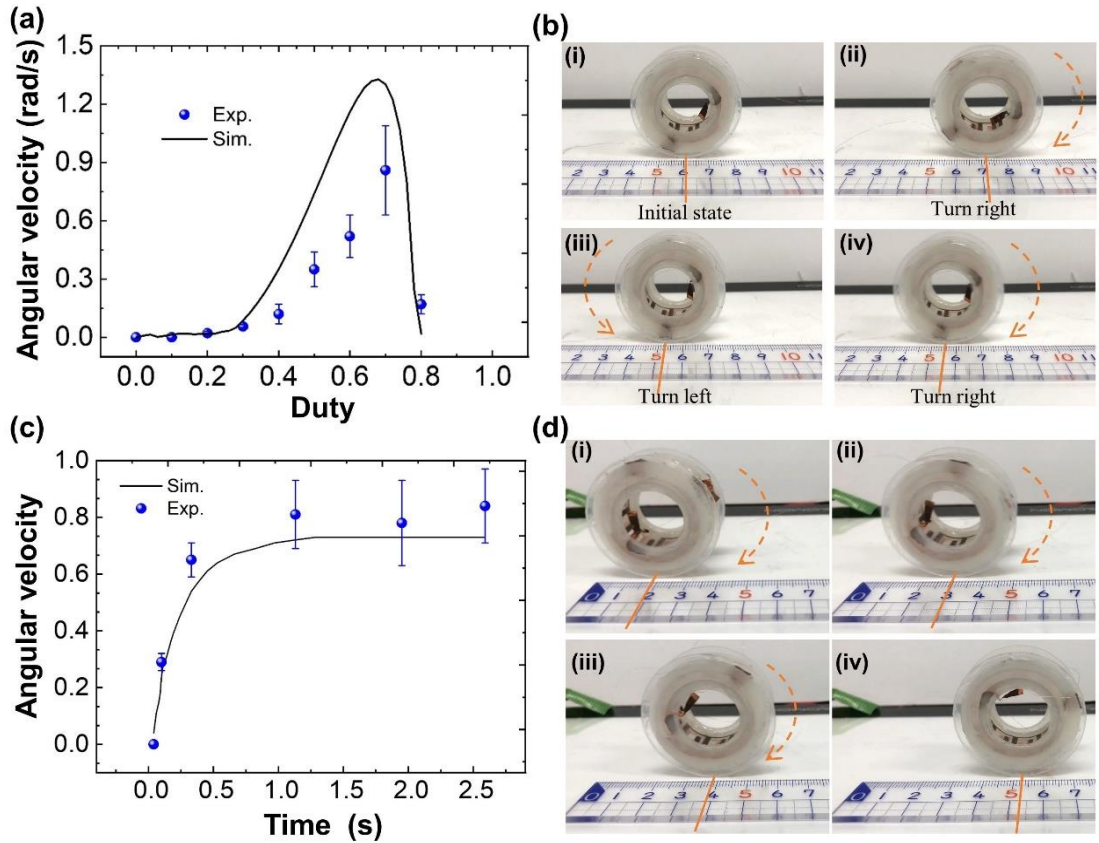

**Fig. 9** Characteristics of the ring-shaped rolling robot. (a) Angular velocity versus duty of applied voltage with rectangular waves. (b) Photos of oscillation motion of ring-shaped robot with its operating condition 4.5 kV (applied voltage), 5 Hz (frequency) and 50% (duty). (c) Photos of rolling motion of our robot with its operating condition 4.5 kV, 5 Hz and 70%.

## 4. Conclusion

In summary, our work encompassed the design, fabrication, and modeling of a wheeled robot actuated by functional fluid. The flowing motion of the EHD fluid has the ability to shift the center point of the ring-shaped robot, thereby inducing rolling locomotion. The EHD pump, which consisted of two layers of PP with copper electrodes, was chosen for its material characteristics, including flexibility and adaptability to bending and twisting. The fabrication of the ring-shaped roller was accomplished using digital manufacturing methods, and we developed both static and dynamic models, incorporating the EHD pump models, to analyze the robot's actuation behavior. Table.1 presents a comparison of soft rolling robots propelled by various types of soft actuators.

Table.1 Comparisons among various rolling robots

| Actuating mechanism | Weight of robots (g) | Average speed (mm/s) | Untethered | Weight and sizes of electrical/light power sources |
|---|---|---|---|---|
| Dielectric elastomer (Li *et al.*, 2021) | 1.02 | ~226 | No | 24.9 kg<br>279×482×654 mm$^3$ |
| Shape memory alloy (Lin, Leisk and Trimmer, 2011) | 16.4 | >50 | No | 9.5 kg<br>213×133×348 mm$^3$ |
| Pneumatic methods (Gong, 2015) | 1469 | ~100 | No | 18.27 kg<br>502×406×406 mm$^3$ |
| Light (Zhang *et al.*, 2014) | ~1 g | ~ 60 | Yes | 2 kg<br>83×120×220 mm$^3$ |
| Magnetic field (Hu *et al.*, 2018) | unknown | 21 | Yes | unknown<br>~ 300×300×350 mm$^3$ |
| Liquid metal (Wu *et al.*, 2018) | 11.27 | ~49 | Yes | ~10 g<br>8×8×14 mm$^3$ |
| Oscillating liquid (Mao *et al.*, 2022) | 1.3 | ~11 | No | 41g<br>62×47×30 mm$^3$ |
| This research | 6.05 | ~205 | No | 43 g<br>58×53×23 mm$^3$ |

The salient contribution of this research lies in its notable achievement of significantly reducing the weight and size of the robots compared to alternative methodologies. In addition, our model demonstrated a strong correlation with the actual experimental results. In the static model, we delved into the influence of several parameters on angular velocity and identified the voltage duty as the core parameter. In the dynamic model, we examined the response of the ring-shaped roller following the application of direct current voltages. Our findings revealed that the wheeled robot can Maintain a relatively consistent pace and attain a peak angular speed of 0.8 radians per second. What sets our approach apart is the absence of intricate moving parts, creating a driving module for this robot that is both lightweight and effortlessly interchangeable, unlike traditional systems that rely on complex components such as motors or shape-memory alloys. In light of these results, we strongly believe that our groundbreaking research has the capacity to accelerate the progress of fluidic actuators powered by flexible EHD pumps.

## Declaration of competing interest

The authors declare that they have no known competing financial interests or personal relationships that could have appeared to influence the work reported in this paper.

## Acknowledgment

This work was supported by Grant-in-Aid for Early-Career Scientists from the Japan Society for the Promotion of Science (Grant No. 23K13290).

## References

Brackenbury, J. (1997) 'Caterpillar kinematics', *Nature,* 390(6659), pp. 453-453.
Bryan, J. and Seyed-Yagoobi, J. (1991) 'Experimental study of ion-drag pumping using various working fluids', *IEEE Transactions on Electrical Insulation,* 26(4), pp. 647-655.
Crossley, V. A. (2006) 'A literature review on the design of spherical rolling robots', *Pittsburgh, Pa,* pp. 1-6.
Ding, R., Wang, J., Cheng, M., Luo, X. and Xu, B. (2024) 'Vibration-polynomial-based optimal trajectory planning for mobile concrete pumping equipment with state constraints', *Automation in Construction,* 158, pp. 105246.
Ding, Z., Sun, Y., Xu, S., Pan, Y., Peng, Y. and Mao, Z. (2023) 'Recent Advances and Perspectives in Deep Learning Techniques for 3D Point Cloud Data Processing', *Robotics,* 12(4), pp. 100.
Fang, Y., Zhang, J., Xu, B., Mao, Z., Li, C., Huang, C., Lyu, F. and Guo, Z. (2021) 'Raising the Speed Limit of Axial Piston Pumps by Optimizing the Suction Duct', *Chinese Journal of Mechanical Engineering,* 34, pp. 1-10.
Felici, N. (1972) 'DC conduction in liquid dielectrics (Part II), electrohydrodynamic phenomena', *Direct current,* 2(4), pp. 147-165.

Floyd, M. and Minor, M. A. 'Impulse based dynamic rolling in the rolling disk biped'. *2010 3rd IEEE RAS & EMBS International Conference on Biomedical Robotics and Biomechatronics*: IEEE, 498-503.

Gong, X. (2015) *Rotary actuators based on pneumatically-driven elastomeric structures.* Rutgers The State University of New Jersey, School of Graduate Studies.

Halme, A., Schonberg, T. and Wang, Y. 'Motion control of a spherical mobile robot'. *Proceedings of 4th IEEE International Workshop on Advanced Motion Control-AMC'96-MIE*: IEEE, 259-264.

Hu, W., Lum, G. Z., Mastrangeli, M. and Sitti, M. (2018) 'Small-scale soft-bodied robot with multimodal locomotion', *Nature,* 554(7690), pp. 81-85.

Huang, X., Kumar, K., Jawed, M. K., Nasab, A. M., Ye, Z., Shan, W. and Majidi, C. (2018) 'Chasing biomimetic locomotion speeds: Creating untethered soft robots with shape memory alloy actuators', *Science Robotics,* 3(25), pp. eaau7557.

Jiao, Z., Ye, Z., Zhu, P., Tang, W., Yang, H. and Zou, J. (2023a) 'Self-sensing actuators with programmable actuation performances for soft robots', *Science China Technological Sciences,* 66(11), pp. 3070-3079.

Jiao, Z., Zhu, P., Hu, Z., Shi, Y., Tang, W., Yang, H. and Zou, J. (2023b) 'Flagellum-Inspired Soft Rotary Motor', *Advanced Materials Technologies,* 8(18), pp. 2300476.

Kong, D., Hirata, T., Li, F., Kurosawa, M. K. and Aoyagi, M. (2024) 'A novel miniature swimmer propelled by 36° Y-cut lithium niobate acoustic propulsion system', *Sensors and Actuators A: Physical,* 365, pp. 114837.

Kong, D., Hirata, T., Wang, Y., Li, F., Kurosawa, M. K. and Aoyagi, M. (2023) 'Acoustic underwater propulsion system based on ultrasonic disc PZT transducer', *Sensors and Actuators A: Physical,* 359, pp. 114502.

Landa, K. and Pilat, A. K. 'Design and start-up of spherical robot with internal pendulum'. *2015 10th International Workshop on Robot Motion and Control (RoMoCo)*: IEEE, 27-32.

Li, W.-B., Zhang, W.-M., Gao, Q.-H., Guo, Q., Wu, S., Zou, H.-X., Peng, Z.-K. and Meng, G. (2021) 'Electrically activated soft robots: Speed up by rolling', *Soft Robotics,* 8(5), pp. 611-624.

Lin, H.-T., Leisk, G. G. and Trimmer, B. (2011) 'GoQBot: a caterpillar-inspired soft-bodied rolling robot', *Bioinspiration & biomimetics,* 6(2), pp. 026007.

Mao, Z., Asai, Y., Yamanoi, A., Seki, Y., Wiranata, A. and Minaminosono, A. (2022) 'Fluidic rolling robot using voltage-driven oscillating liquid', *Smart Materials and Structures,* 31(10), pp. 105006.

Mao, Z., Iizuka, T. and Maeda, S. (2021) 'Bidirectional electrohydrodynamic pump with high symmetrical performance and its application to a tube actuator', *Sensors and Actuators A: Physical,* 332, pp. 113168.

Mao, Z., Nagaoka, T., Yokota, S. and Kim, J.-w. (2020) 'Soft fiber-reinforced bending finger with three chambers actuated by ECF (electro-conjugate fluid) pumps', *Sensors and Actuators A: Physical,* 310, pp. 112034.

Mao, Z., Peng, Y., Hu, C., Ding, R., Yamada, Y. and Maeda, S. (2023) 'Soft computing-based predictive modeling of flexible electrohydrodynamic pumps', *Biomimetic Intelligence and Robotics,* 3(3), pp. 100114.

Mao, Z., Yoshida, K. and Kim, J.-w. (2019a) 'Developing O/O (oil-in-oil) droplet generators on a chip by using ECF (electro-conjugate fluid) micropumps', *Sensors and Actuators B: Chemical,* 296, pp. 126669.

Mao, Z., Yoshida, K. and Kim, J.-w. (2019b) 'A droplet-generator-on-a-chip actuated by ECF (electro-conjugate fluid) micropumps', *Microfluidics and Nanofluidics,* 23(12), pp. 130.

Matsubara, T., Choi, J. S., Kim, D. H. and Kim, J. w. (2022) 'A Microfabricated Pistonless Syringe Pump Driven by Electro-Conjugate Fluid with Leakless On/Off Microvalves', *Small,* 18(15), pp. 2106221.

Ng, C. S. X., Tan, M. W. M., Xu, C., Yang, Z., Lee, P. S. and Lum, G. Z. (2021) 'Locomotion of miniature soft robots', *Advanced Materials,* 33(19), pp. 2003558.

Peng, Y., Li, D., Yang, X., Ma, Z. and Mao, Z. (2023a) 'A review on electrohydrodynamic (EHD) pump', *Micromachines,* 14(2), pp. 321.

Peng, Y., Nabae, H., Funabora, Y. and Suzumori, K. (2024) 'Controlling a peristaltic robot inspired by inchworms', *Biomimetic Intelligence and Robotics,* 4(1), pp. 100146.

Peng, Y., Sakai, Y., Nakagawa, K., Funabora, Y., Aoyama, T., Yokoe, K. and Doki, S. (2023b) 'Funabot-Suit: A bio-inspired and McKibben muscle-actuated suit for natural kinesthetic perception', *Biomimetic Intelligence and Robotics,* 3(4), pp. 100127.

Potz, M., Artusi, M., Soleimani, M., Menon, C., Cocuzza, S. and Debei, S. (2010) 'Rolling dielectric elastomer actuator with bulged cylindrical shape', *Smart materials and structures,* 19(12), pp. 127001.

Preston, D. J., Jiang, H. J., Sanchez, V., Rothemund, P., Rawson, J., Nemitz, M. P., Lee, W.-K., Suo, Z., Walsh, C. J. and Whitesides, G. M. (2019) 'A soft ring oscillator', *Science Robotics,* 4(31), pp. eaaw5496.

Robertson, M. A. and Paik, J. (2017) 'New soft robots really suck: Vacuum-powered systems empower diverse capabilities', *Science Robotics,* 2(9), pp. eaan6357.

Sugiyama, Y., Shiotsu, A., Yamanaka, M. and Hirai, S. 'Circular/spherical robots for crawling and jumping'. *Proceedings of the 2005 IEEE international conference on robotics and automation*: IEEE, 3595-3600.

Wu, J., Tang, S. Y., Fang, T., Li, W., Li, X. and Zhang, S. (2018) 'A wheeled robot driven by a liquid-metal droplet', *Advanced materials,* 30(51), pp. 1805039.

Yamada, M., Kondo, M., Mamiya, J. i., Yu, Y., Kinoshita, M., Barrett, C. J. and Ikeda, T. (2008) 'Photomobile polymer materials: towards light-driven plastic motors', *Angewandte Chemie International Edition,* 47(27), pp. 4986-4988.

Zhang, C., Chen, J., Li, J., Peng, Y. and Mao, Z. (2023) 'Large language models for human-robot interaction: A review.', *Biomimetic Intelligence and Robotics*.

Zhang, X., Yu, Z., Wang, C., Zarrouk, D., Seo, J.-W. T., Cheng, J. C., Buchan, A. D., Takei, K., Zhao, Y. and Ager, J. W. (2014) 'Photoactuators and motors based on carbon nanotubes with selective chirality distributions', *Nature communications,* 5(1), pp. 2983.